\documentclass{article}

\usepackage{PRIMEarxiv}
\pdfoutput=1
\usepackage[utf8]{inputenc} 
\usepackage[T1]{fontenc}    
\usepackage{hyperref}       
\usepackage{url}            
\usepackage{booktabs}       
\usepackage{amsfonts}       
\usepackage{nicefrac}       
\usepackage{microtype}      
\usepackage{lipsum}
\usepackage{fancyhdr}       
\usepackage{graphicx}       
\usepackage{apacite}
\usepackage{float}
\usepackage{caption}
\usepackage{parskip}
\usepackage{fancyref}
\usepackage{subcaption}
\usepackage{float}
\usepackage{amsmath}
\usepackage{cleveref}
\graphicspath{{media/}}     
\title{Risk-averse Batch Active Inverse Reward Design
}

\author{
  Panagiotis Liampas
  \thanks{Guided and advised by Peter McIntyre (Non-Trivial)}\\
  \texttt{pliam1105@gmail.com} \\
}

\begin{document}
\maketitle

\begin{abstract}
Designing a perfect reward function that depicts all the aspects of the intended behavior is almost impossible, especially when having to deal with generalizing it outside of the training environments. Active Inverse Reward Design (AIRD) proposed the use of a series of queries, comparing possible reward functions in a single training environment. This allows the human to give information to the agent about suboptimal behaviors, in order to compute a probability distribution over the intended reward function. However, it ignores the possibility of unknown features appearing in real-world environments, and the safety measures needed until the agent completely learns the reward function. I improved this method and created Risk-averse Batch Active Inverse Reward Design (RBAIRD), which constructs batches, sets of environments the agent encounters when being used in the real world, processes them sequentially, and, for a predetermined number of iterations, asks queries that the human needs to answer for each environment of the batch. After this process is completed in one batch, the probabilities have been improved and are transferred to the next batch. This makes it capable of adapting to real-world scenarios and learning how to treat unknown features it encounters for the first time. I also integrated a risk-averse planner, similar to that of Inverse Reward Design (IRD), which samples a set of reward functions from the probability distribution and computes a trajectory that takes the most certain rewards possible. This ensures safety while the agent is still learning the reward function, and enables the use of this approach in situations where cautiousness is vital. RBAIRD outperformed the previous approaches in terms of efficiency, accuracy, and action certainty, demonstrated quick adaptability to new, unknown features, and can be more widely used for the alignment of crucial AI models that have the power to significantly affect our world.
\end{abstract}

\section{Introduction}
\label{sec:sec_intro}
AI Safety is becoming increasingly important in recent years, due to the immense development of the tools used to design, train, and deploy AI models, that are accessible to the public. This development is \textbf{largely disproportional} to the effort put into finding safe approaches for training an AI to be compliant with human values and restricted to its specific task. According to \shortcite{hiltonPreventingAIrelatedCatastrophe2023}, only approximately 400 people are working in that field. According to the same source, only 50 million dollars are put into AI safety compared to the 1 billion dollars used for accelerating AI development.\par
\begin{figure}[ht]
  \begin{subfigure}[t]{0.37\textwidth}
    \centering
    \includegraphics[width=\textwidth]{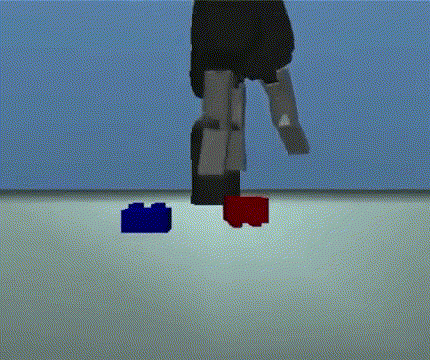}
    \caption{An example of reward hacking: instead of the robot stacking the red block on top of the blue one, it flips it and takes the reward, which was specified using the height of the base of the block \shortcite{popovDataefficientDeepReinforcement2017}.}
    \label{fig:img_reward_misspec}
  \end{subfigure}
  \hfill
  \begin{subfigure}[t]{0.57\textwidth}
    \centering
    \includegraphics[width=\textwidth]{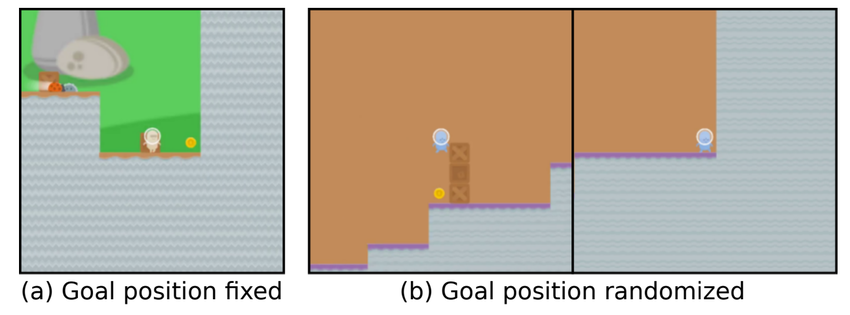}
    \caption{A demonstration of goal misgeneralization: the agent was only trained in environments where the goal was at the end of the level so, during testing, it continues going right even if the goal is in the middle \shortcite{langoscoGoalMisgeneralizationDeep2023}.}
    \label{fig:img_goal_misgen}
  \end{subfigure}
  \caption{Overview of two basic issues in Reinforcement Learning, which cause a lot of problems when trying to align the agent with humans' goals.}
  \label{fig:img_rl_issues}
\end{figure}

Reward Design is the problem of finding the appropriate reward function that causes the desired \textbf{behavior} of the agent\footnote{As an \textit{agent} we define every model that is deployed into an environment to perform a specific task} in all the environments where it is deployed. The important point is that we don't just want the desired \textit{trajectory} in a training environment, but a \textit{policy} that doesn't cause the agent to pursue power or use manipulative methods, \textbf{``hack the reward"} \shortcite{panEffectsRewardMisspecification2022}, like in \Fref{fig:img_reward_misspec}, to achieve his goal. This is very closely related to \textbf{goal misgeneralization}, shown in \Fref{fig:img_goal_misgen}, and is caused by our inability to create training environments that demonstrate the intended policy in all the possible situations that the agent will encounter in the real world \shortcite{langoscoGoalMisgeneralizationDeep2023}.\par
There have been remarkable efforts to solve that particular problem, especially in the field of Reinforcement Learning from Human Feedback \shortcite{casperOpenProblemsFundamental2023}. One of them is Active Inverse Reward Design (AIRD) \shortcite{mindermannActiveInverseReward2019}, which uses the capability enabled by Inverse Reward Design (IRD) \shortcite{hadfield-menellInverseRewardDesign2020}, to compute a \textbf{probability distribution} over the true reward function based on a \textbf{human-made} estimation and a training environment, and enhances it with \textbf{human queries} and the use of comparisons between \textbf{suboptimal behaviors} to infer the required policy. However, both of these methods are still applied in a single training environment, not being able to adapt to \textbf{new features} present in the real world, therefore not capturing the behavior completely. Also, considering only the features present in that environment, it usually doesn't offer the opportunity to depict that behavior in \textbf{all possible situations}, even by considering suboptimal policies. Finally, even when the agent becomes completely certain about the intended behavior, that doesn't happen immediately, and it is crucial to ensure that, in the first iterations of the process, it only follows steps where it is the most \textbf{certain they are safe}, in a trajectory that is applied to the real world.\par
My work builds upon the existing approaches and focuses on tackling their limitations, which I listed above. Something that we can infer from the first two issues is that we want to be able to \textbf{update the agent's beliefs} about the reward function in the \textbf{test environments} and repeat the querying process in them as well. Doing that in every test environment (real-world use case) is very inefficient and requires a large amount of human input and time. Therefore, the first key observation is to create \textbf{batches of test data}, where each one contains lots of them, and apply AIRD to them.\par
The first capability that this change adds is \textbf{adapting to new environments}, with features never seen before, therefore tackling a large part of the goal misgeneralization problem. It also allows the agent to learn \textbf{various aspects of the behaviors} caused by the queried reward functions, by applying them to different environments instead of just one, largely increasing the \textbf{information gain} of a single query.\par
The safety issue mentioned above leads to the second observation, which is that we can use a \textbf{risk-averse planning method}, similar to that used in IRD or some other more efficient one, to make the agent greatly value \textbf{certainty} instead of only the reward that it estimates it will get from the reward distribution computed.\par
These two observations comprise the basic structure of \textbf{Risk-averse Batch Active Inverse Reward Design (RBAIRD)}. The steps involved in the process of implementing that method and measuring its improvement over the previous ones were the following: 1) Creating batches and applying the querying process to them; 2) Integrating some simple risk-averse planning methods; 3) Performing experiments where I vary the number of batches, their size, and the risk-averse method used; 4) Performing experiments where I add new features in each batch, to evaluate its ability to adapt to new environments.
\section{Background}
\begin{figure}[tb]
  \begin{subfigure}[t]{0.47\textwidth}
    \centering
    \includegraphics[width=\textwidth]{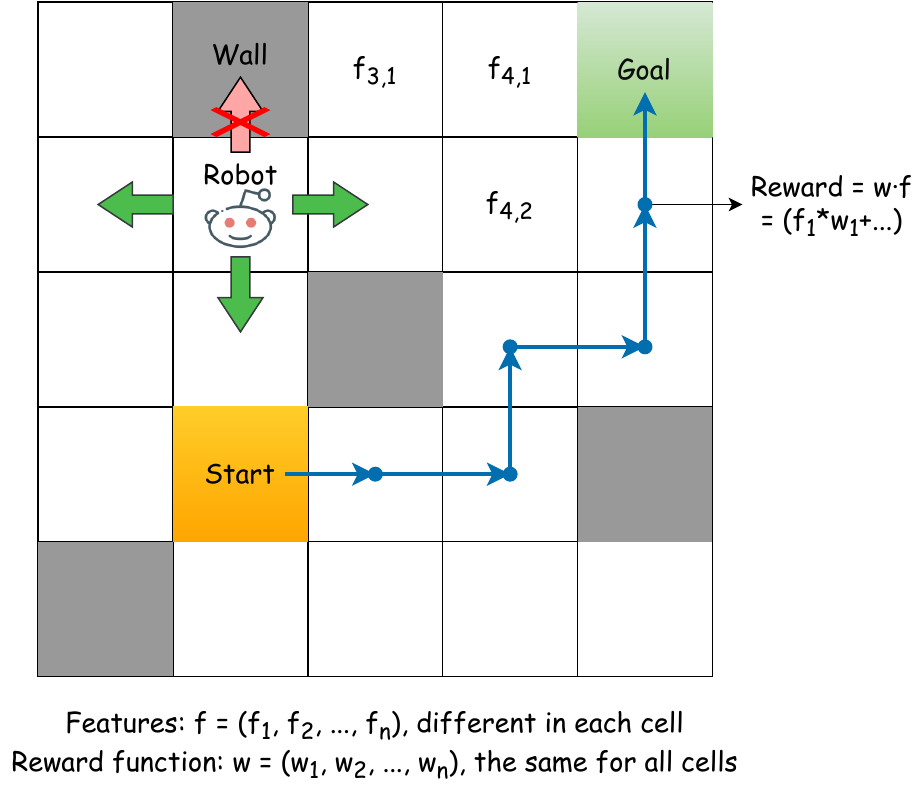}
    \caption{The environment where the MDP optimized by the agent was specified. The goal of the agent was to maximize the sum of the rewards of the cells in the trajectory.}
    \label{fig:fig_env}
  \end{subfigure}
  \hfill
  \begin{subfigure}[t]{0.47\textwidth}
    \centering
    \includegraphics[width=\textwidth]{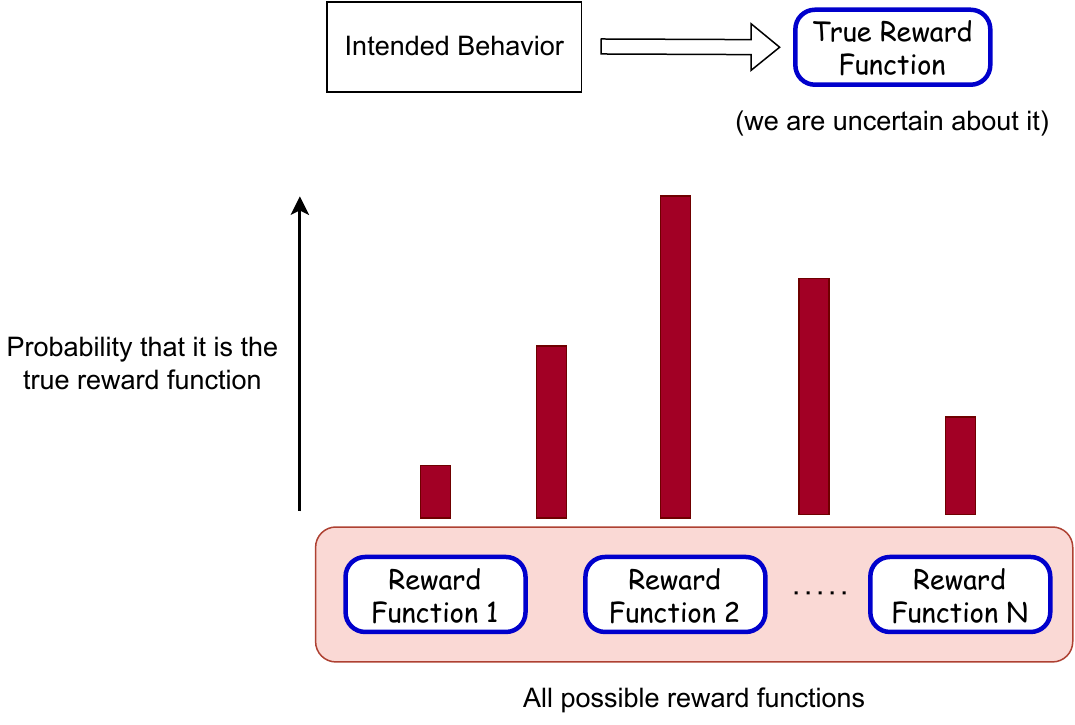}
    \caption{We compute the probability distribution over the true reward function, which is the agent's belief about the intended behavior, including its uncertainty.}
    \label{fig:fig_prob_distrib}
  \end{subfigure}
  \caption{The core elements of IRD, AIRD, and RBAIRD: the environment used and the probability distribution that we want to compute in order to find the desired reward function.}
  \label{fig:fig_setup}
\end{figure}

\paragraph{Environment.} The environment that I used is a \textit{gridworld} (see \Fref{fig:fig_env}), which is a grid with dimensions $12\times12$, and it contains:
\begin{itemize}
  \item A \textit{robot}, which can move up, down, right, and left in adjacent cells.
  \item A \textit{start} state, from which the robot starts moving.
  \item Some \textit{goal} states, which when the robot reaches it stops moving.
  \item Some \textit{walls}, from which the robot cannot pass through.
  \item All the other cells, in which the robot moves.
\end{itemize}\par
All the cells contain a vector of \textit{features} $(f_1, f_2, \ldots, f_n)$, which are used to calculate the reward in that state. The reward is computed using a \textit{reward function}, which is the same vector of weights $(w_1, w_2, \ldots, w_n)$ along all states. The reward in a state with features $f=(f_1, f_2, \ldots, f_n)$ and weights $w=(w_1, w_2, \ldots, w_n)$ is their dot product $f\cdot w=(f_1\cdot w_1+f_2\cdot w_2+\ldots+f_n\cdot w_n)$. From this product, I also subtract a \textit{living reward}, that is used to incentivize shorter routes.\par
A \textit{policy} is a map from the states $(x,y)$ to the \textit{action} (north, south, east, west) in the environment. An \textit{agent} controls the robot and moves it in specific directions, using a predetermined policy, in order to maximize the total reward in the trajectory of the robot (the \textit{trajectory} is the set of states the robot has visited in chronological order until we stopped it or it reached a goal).\par
\paragraph{Finding the true reward function.}
\label{sec:par_finding_true_reward_function}
In both IRD, AIRD and my approach, we try to find the reward function that best represents the intended behavior of the agent, which we call the \textit{true reward function}. This function is an element of a big set, the \textit{true reward space}, which contains all the possible reward functions.\par
However, because we are unsure of that perfect reward function, in IRD we start with a human-made estimate which is the \textit{proxy reward function}, an element of the \textit{proxy reward space} (as in AIRD we start with no information about the reward function, we are only given that space). The goal of the previous mentioned papers and my approach is to find a \textbf{probability distribution} over all the rewards in the true reward space: for each element of it, we want the probability that it is the true reward function, based on the behavior they incentivize in the training environment, as shown in \Fref{fig:fig_prob_distrib}.\par
The \textit{feature expectations}, given a reward function and an environment, are the expected sum of the features in a trajectory derived from an optimal policy. In both the total trajectory reward and the feature expectations, we apply a discount $\gamma$ (it might be $1$), such that the next feature or reward is first multiplied by $\gamma^i$, where $i$ increases by $1$ each time the robot moves.
\subsection{Inverse Reward Design}
\begin{figure}[tb]
  \centering
  \includegraphics[width=\textwidth]{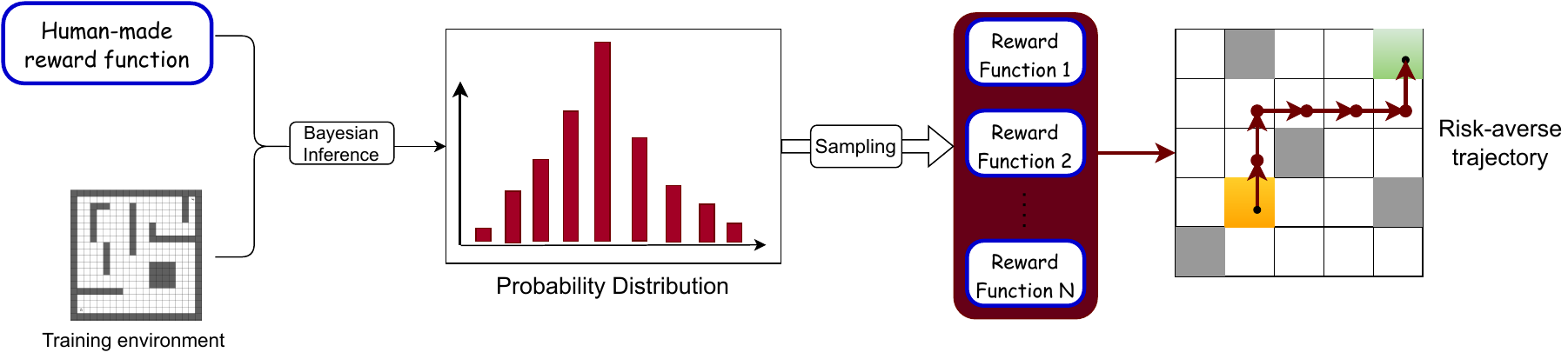}
  \caption{An overview of the IRD approach: given an environment and a human estimate of the intended reward function, it computes for each reward function the probability that it is the intended one, and uses that to determine a policy that avoids uncertain actions.}
  \label{fig:fig_ird}
\end{figure}

In Inverse Reward Design (IRD) \shortcite{hadfield-menellInverseRewardDesign2020}, given the true reward function and the proxy reward function, they use Bayesian inference to compute the probability distribution over the \textbf{true reward function}, as demonstrated in \Fref{fig:fig_ird}.\par
It then computes a \textbf{risk-averse policy}, that takes actions so that the distribution of the rewards in that state, found using a set of weights sampled with the precomputed probabilities and the features of that state, has low variance. The risk-averse policy can be computed in various ways, like by maximizing the worst-case reward, per state or trajectory, or comparing the reward of each state with the reward of some baseline features used as a reference point.
\subsection{Active Inverse Reward Design}
\begin{figure}[tb]
  \centering
  \includegraphics[width=0.618\textwidth]{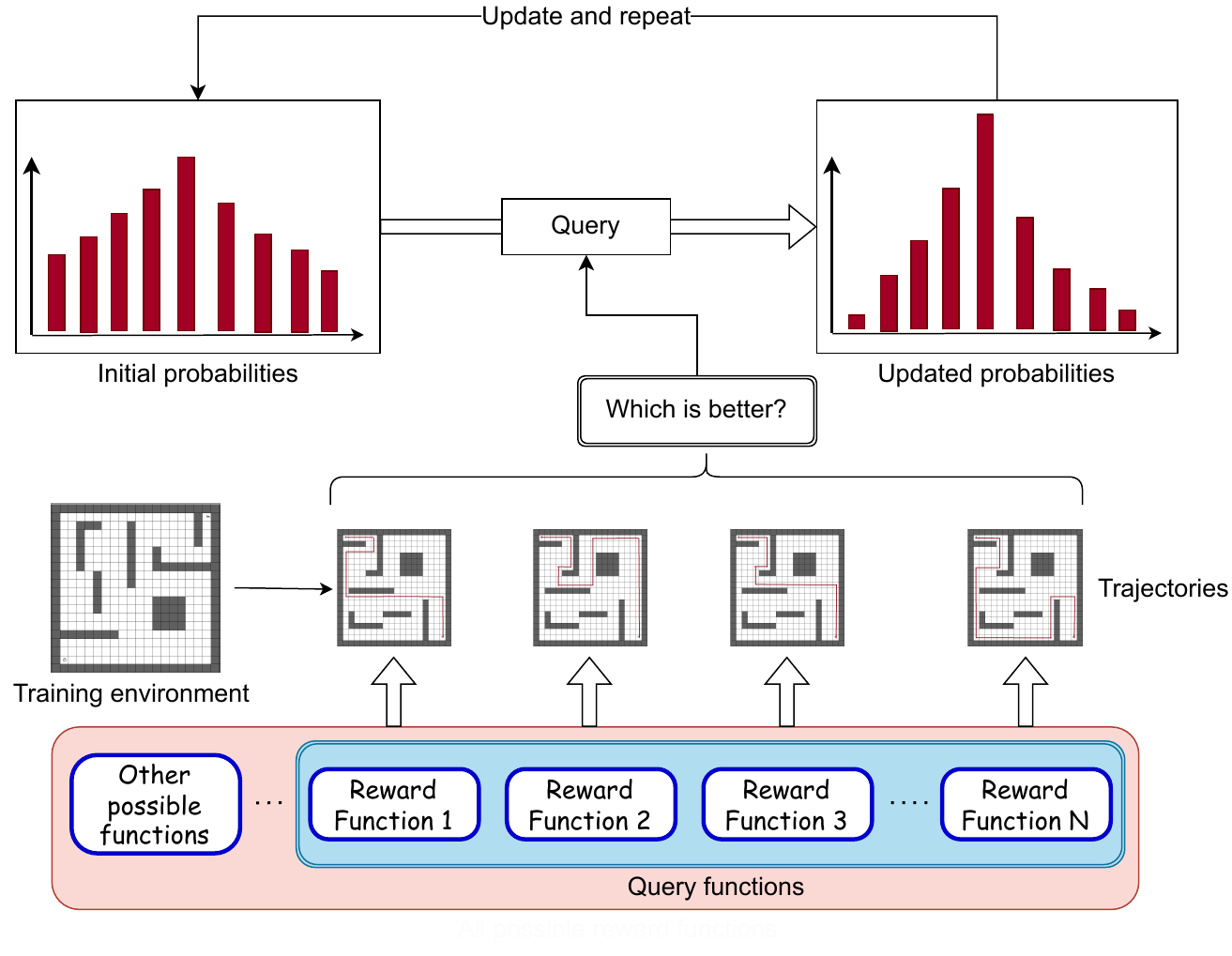}
  \caption{In AIRD, human feedback on the reward functions is used to find the intended one, by iteratively comparing suboptimal reward functions and choosing which one better performs in the training environment.}
  \label{fig:fig_aird}
\end{figure}

In Active Inverse Reward Design (AIRD) \shortcite{mindermannActiveInverseReward2019}, given the true reward space and a proxy reward space, they continuously ask \textbf{human queries} in order to update the desired probability distribution (starting with the uniform distribution as we don't know anything about it), as shown in \Fref{fig:fig_aird}. Their approach is based on the capability, which IRD introduced, to infer that distribution using a faulty approximation of the reward function.
\paragraph{Queries.} A \textit{discrete query} is defined as a subset of the proxy reward space, whose answer is an element of that subset that the human thinks best \textbf{incentivizes the desired behavior}, compared to the other elements of that subset. There is also the ability to use \textit{feature queries}, which ask the human to set specific variable feature weights of a reward function, while the other weights are fixed. However, in my work, I only examined the first type of queries, due to performance issues. It is important to note that this approach utilizes the information gained from \textbf{suboptimal} behaviors to form a more accurate and well-rounded belief about the reward function.
\paragraph{Updating probabilities.} After each query, it uses Bayesian inference to \textbf{update the probability distribution} based on the answer to that query. To do that, it uses a Q learning planner that optimizes trajectories in the training environment shown in \Fref{fig:fig_env}, setting each element of the query as the reward function. It then computes the feature expectations of these trajectories and uses these and the query answer to update the probabilities.
\paragraph{Query selection.}
\label{sec:par_query_selection_aird}
The queries are chosen greedily, such that the \textbf{expected information gain} of the answer to that is maximal. The information gain is measured using various metrics, one of which is the entropy of the probability distribution over the true reward function. There are multiple ways this selection can be done, as described in the original paper, but the one I used for my approach due to its efficiency is the following: as long as the query size is less than a predetermined constant, we take a random vector of weights and maximize the information gain when these weights are the query answer, by taking gradient descent steps on them.
\section{Batches of Environments}
The key aspect of RBAIRD that improves upon AIRD is the fact that it uses multiple environments instead of one, in order to capture different aspects of the same reward function. A way this could be done is by applying the AIRD process in \textit{real time}, in each environment that the agent is deployed on, e.g., in the real world.
However, this is very inefficient, as the processes of selecting the queries and training the agent on the new distribution is computationally expensive.\par
\begin{figure}[tb]
  \centering
  \includegraphics[width=\textwidth]{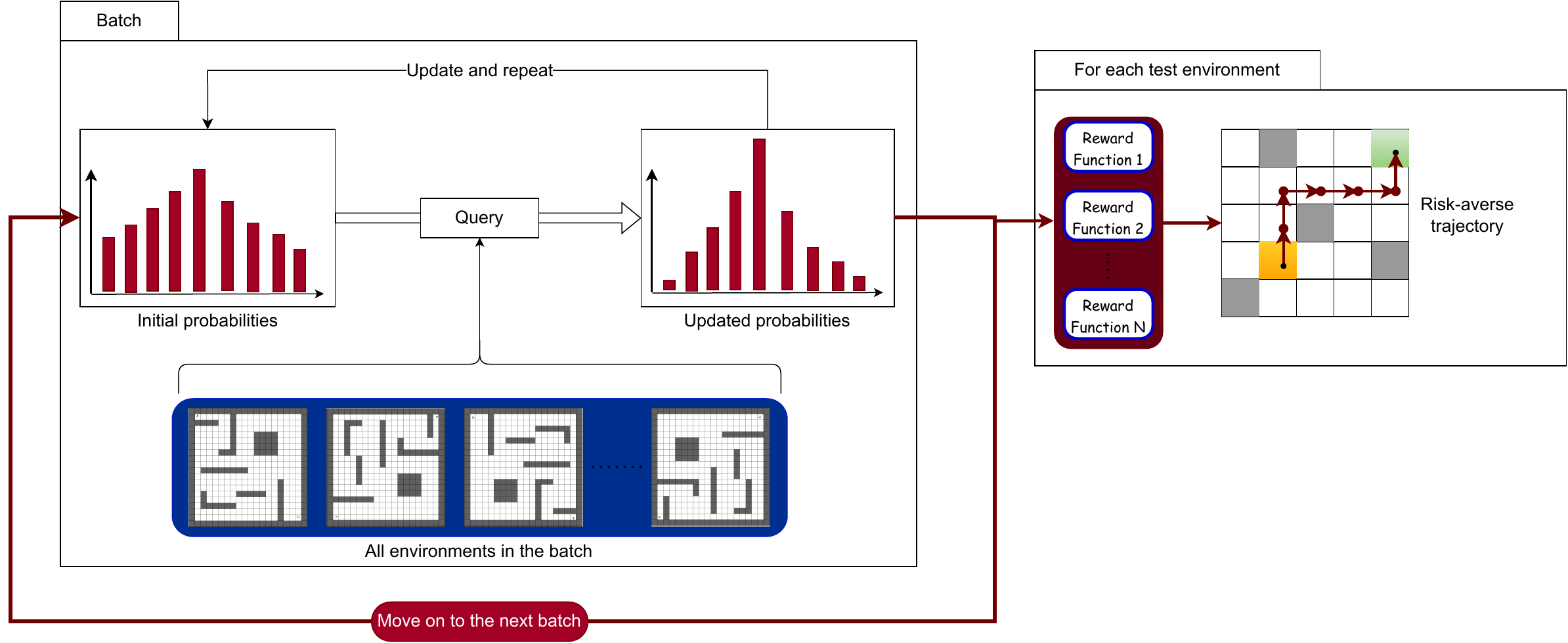}
  \caption{An overview of the RBAIRD process: For each batch, we iterate over the process of selecting a query and answering it for each environment. We then train the agent to adopt a risk-averse policy that values the certainty of the rewards received in real-world environments.}
  \label{fig:fig_rbaird_overview}
\end{figure}

\paragraph{Batches.} On the other hand, by separating the environments into \textbf{batches}, which are defined as subsets with a specific number of environments in each one, and applying the query process in all environments of each batch at once, as demonstrated in \Fref{fig:fig_rbaird_overview}, we can increase the \textbf{time efficiency} of the process, and have a bigger \textbf{information gain} for each query, as it captures the behavior of a reward function in all these environments. As the agent is deployed in \textbf{real-world scenarios}, using a certain initial probability distribution, it would store the environments encountered and add them to the current batch, until we reach the desired size. It would then \textbf{update the distribution}, by applying the query process in the batch, and train on the new distribution. This way, it would be refined using real-world data, while generalizing on them instead of deriving its policy only from training.
\paragraph{Query selection.} The querying process is largely similar to that of AIRD, described in \nameref{sec:par_query_selection_aird}, but adapted to \textbf{multiple environments}. Specifically, for each batch, in each iteration of the process, I select a single query that has the \textbf{maximum information gain} when answered to all the environments in the batch. The way we compute that gain is by using the current probabilities to find the expected answer to the query, and applying it using Bayesian inference to compute the entropy of the updated distribution (see \nameref{sec:par_inference}). The information gain is measured as the difference between the new entropy and the previous one (which used the initial probabilities).
\paragraph{Query answering.} The way a query is answered in the batch is by selecting, for each environment in it, the reward function that best performs in that specific environment, as shown in \Fref{fig:fig_rbaird_queries}. It should be noted that, while we select the \textbf{same query} for all environments of the batch, for increased efficiency of the process, that query is answered in each environment \textbf{separately}. This demonstrated \textbf{more accurate results} than answering each query at once for all environments, due to the inability of a reward function to be comparatively optimal in all environments, as each of them can focus on a different aspect of that function. It is also \textbf{clearer to answer} for a human, which can judge a function mostly by its result in the environment used, which can be different in each one, and therefore a single answer for all environments would be ambiguous.\par
\begin{figure}[tb]
  \centering
  \includegraphics[width=\textwidth]{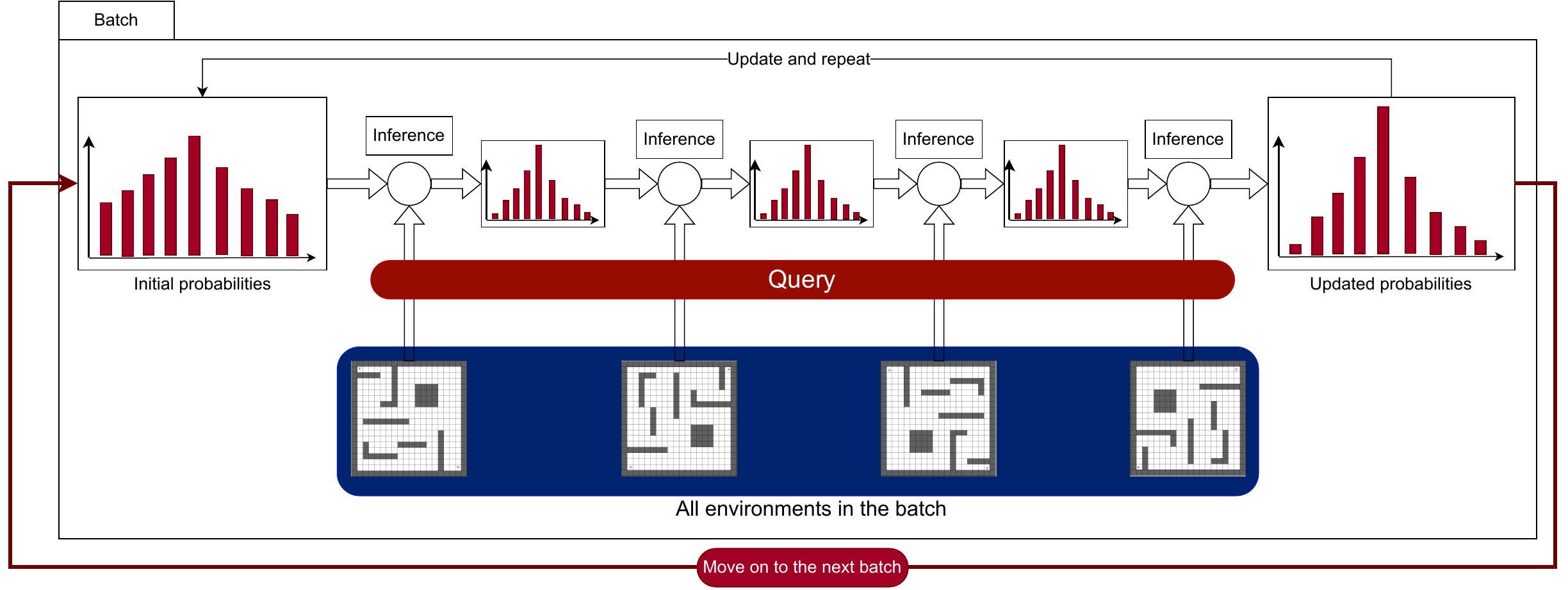}
  \caption{A diagram that shows how the query answers from all environments are combined in order to update the probability distribution over the true reward function, by taking the environments one by one and each time updating the probabilities using the answer for that environment.}
  \label{fig:fig_rbaird_batches}
\end{figure}

\paragraph{Inference.}
\label{sec:par_inference}
Both for the query selection process and the refinement of the agent's belief about the reward function, we consider the environments of the batch \textbf{sequentially}, as shown in \Fref{fig:fig_rbaird_batches}. For each environment, we compute the \textbf{feature expectations}, as described in \nameref{sec:par_finding_true_reward_function}, using the answer of the query for it (or, when selecting the query, the expected answer based on the current belief), as well as the other functions in the query. We then apply Bayesian inference to \textbf{update the probability distribution} over the true reward function, in a way similar to AIRD \shortcite{mindermannActiveInverseReward2019}. Therefore, when we move on to the next environment in the batch, we start with the distribution that was updated using the previous environment. When we finish with the last environment as well, we have the final updated probabilities over \textbf{all the environments}, using the query. Also, when we finish with a batch (after some iterations of the above process, with a new query selected in each iteration), the final updated probabilities are \textbf{passed on} to the next batch as the initial probabilities. This way, all the information gained from a batch is then able to be used in a real-life scenario, after \textbf{acquiring knowledge} about a potentially unknown and unsafe environment.
\section{Risk-averse Planning}
\label{sec:sec_risk_averse}
\begin{figure}[tb]
  \begin{subfigure}[t]{0.47\textwidth}
    \centering
    \includegraphics[width=\textwidth]{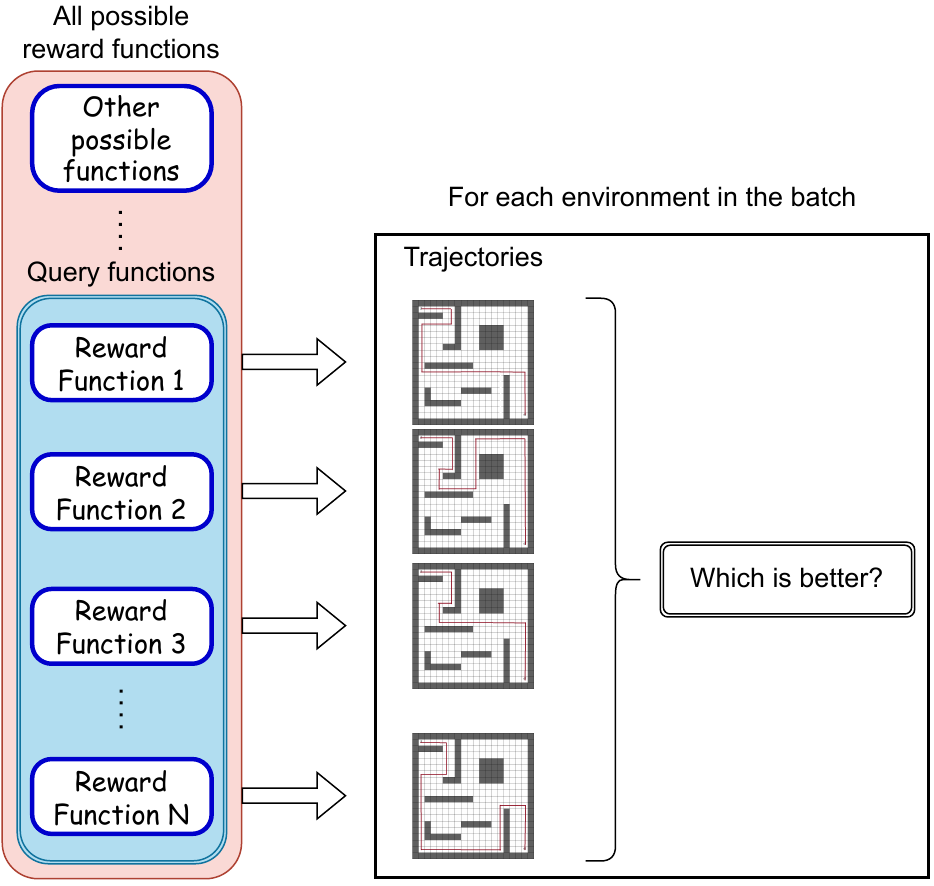}
    \caption{An overview of how queries are defined in RBAIRD: from a subset of all the possible reward functions, for each environment in the batch, the human chooses which one demonstrates the best policy, compared to the others, in that environment.}
    \label{fig:fig_rbaird_queries}
  \end{subfigure}
  \hfill
  \begin{subfigure}[t]{0.47\textwidth}
    \centering
    \includegraphics[width=\textwidth]{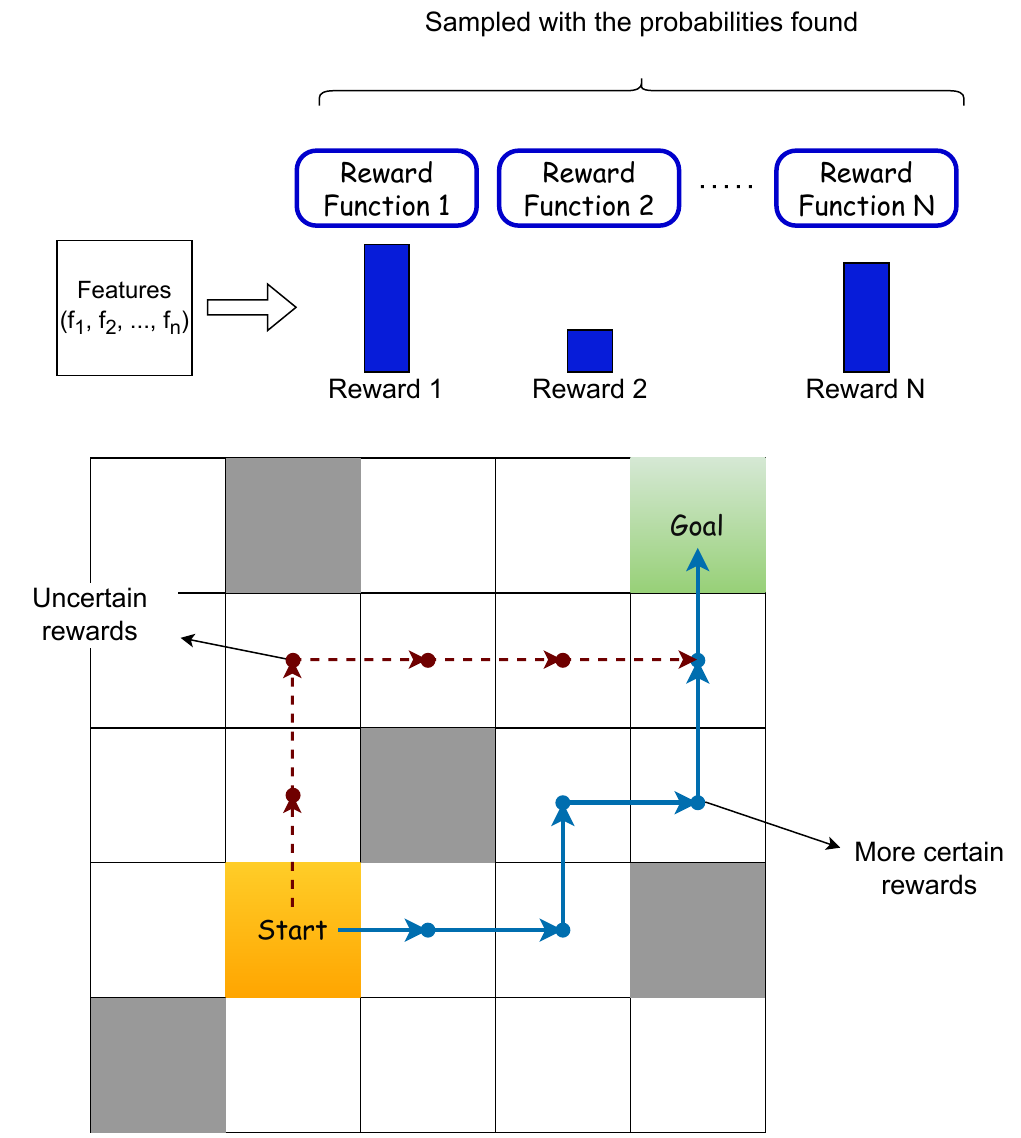}
    \caption{How risk-averse planning works: we sample a set of weights from the probability distribution we have been updating with the queries, compute the set of rewards of the current state based on these weights, and penalize the agent if that set has high variance.}
    \label{fig:fig_rbaird_risk_averse}
  \end{subfigure}
  \caption{Diagrams explaining two key aspects of the RBAIRD approach: the new form of queries adapted to multiple environments, and the risk-averse planning method that ensures the agent takes safe actions.}
  \label{fig:fig_queries_risk_averse}
\end{figure}

As mentioned in \nameref{sec:sec_intro}, deploying an agent that progressively learns the intended behavior, using the method described above, means that during the period that it is uncertain about it, it might demonstrate \textbf{unwanted} and unpredictable behavior that can cause serious damage if given enough power. The way I tried to tackle this issue is by utilizing a \textbf{risk-averse planning} method similar to that of IRD \shortcite{hadfield-menellInverseRewardDesign2020}, which incentivizes the agent to take the most certain actions possible, meaning the ones where the reward of the resulting state (or the entire trajectory) has the \textbf{minimum variance} possible, while still getting big rewards. I made a planner that uses the Q-Learning algorithm \shortcite{watkinsQlearning1992} to compute the optimal policy, and then the required trajectory, of an agent that receives a per-state reward that I designed to incentivize low variance actions, as shown in \Fref{fig:fig_rbaird_risk_averse}.\par
To determine that reward, the probability distribution over the true reward function is used to sample a set of weights. For each state, I computed the set of rewards produced by multiplying the state features with the weight vectors sampled by the distribution. Then, the reward for that state was determined in one of the following two ways (chosen for each experiment):
\begin{itemize}
  \item By taking the \textbf{minimum (worst-case) reward} from the reward set of that state.
  \item By \textbf{subtracting the variance} of the reward set, multiplied by a predetermined coefficient, from the average of that set (the expected reward).
\end{itemize}
These simple methods achieve the expected result by reducing the variance of the trajectories planned, but more efficient methods are listed in the IRD paper, while other sophisticated and complex methods can reduce the observed phenomenon of \textit{blindness to success}, as discussed in \nameref{sec:sec_lim_future_work}.
\section{Evaluation}
\label{sec:sec_eval}
An important part of my work was measuring the \textbf{performance} of my approach, especially how \textbf{accurate} the computed probability distribution was regarding the true reward function, and how great the risk-averse planner was at \textbf{reducing the variance} of the trajectories, as well as its deviation in performance from the optimal one due to it. Also, it was very crucial, as it was a key goal of this approach, to see how \textbf{fast} it finds the true reward function, and how well it adapts to \textbf{new environments}.
\paragraph{Planners.} To evaluate my approach, I needed to compare its performance with the one without risk-averse planning, so I made an \textbf{unsafe planner} that only cared about the \textbf{expected reward}. To do that, I used Q Learning in the same way as before, but the per-state reward was the \textbf{average} of the set of rewards computed by multiplying the sample of weights from the probability distribution with the feature state vector. I also made an \textbf{optimal planner} that works in the same way, but, instead of using a set of rewards, it is given the exact true reward function as an input, acting as the \textbf{baseline} for understanding how close the agent is to finding that function, and comparing the policies of the unsafe and the risk-averse planner with the optimal performance.
\paragraph{Test environments.}
\label{sec:par_test_envs}
As the number of environments in each batch could be small, depending on the experiment, but I needed \textbf{high accuracy} on the average performance of all planners (the safe, unsafe, and the optimal one), I created a random \textbf{big set} of test environments, that were the same for all batches. After updating the agent on each batch, I used the planners, given the updated probability distribution, to compute the \textbf{expected trajectories} in each test environment and the respective feature expectations. Then, I used them to compute the metrics described in \nameref{sec:par_metrics}.
\begin{figure}[tb]
  \centering
  \includegraphics[width=0.8\textwidth]{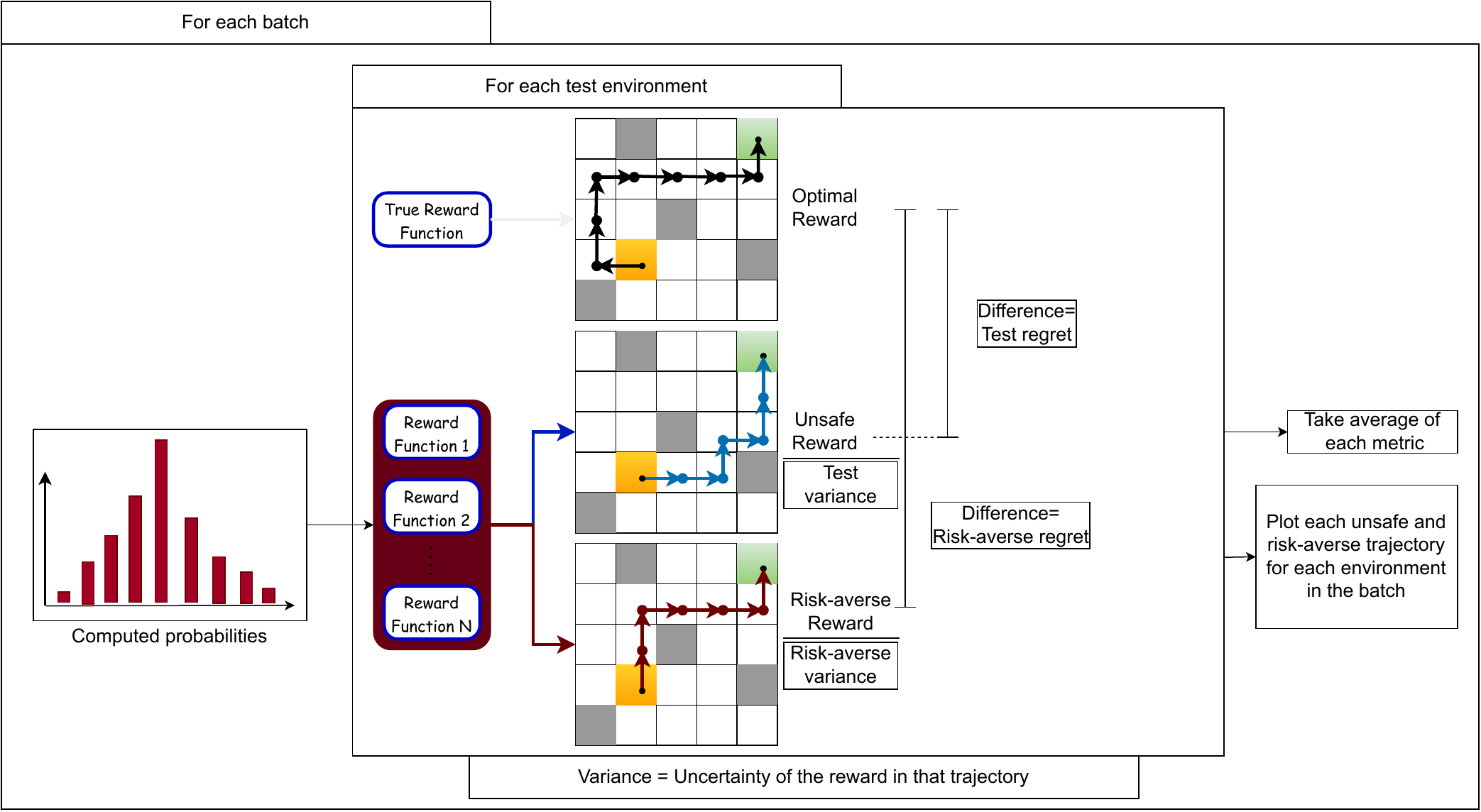}
  \caption{An overview of the evaluation process: after finishing the querying and training process for each batch, I compute the optimal policy using a planner that knows the true reward function, and two suboptimal policies using an unsafe and a risk-averse planner given the computed probabilities. Then, I compute various metrics that measure the performance and accuracy of the process, as well as the certainty of the actions of the agents.}
  \label{fig:fig_rbaird_eval}
\end{figure}
\paragraph{Metrics.}
\label{sec:par_metrics}
I multiplied the feature expectations of all planners, computed as described in \nameref{sec:par_test_envs}, with the true reward function (as this is what matters in real-world scenarios) and got their policies' \textbf{true reward}, which we call \textit{optimal reward}, \textit{unsafe reward}, and \textit{risk-averse reward}, depending on the planner used. Based on these, we compute the $\textit{test regret} = \textit{optimal reward} - \textit{unsafe reward}$, and the $\textit{risk-averse regret} = \textit{optimal reward} - \textit{risk-averse reward}$. Specifically, \textit{test regret} measures how \textbf{accurate} the probability distribution is relative to the true reward function, and \textit{risk-averse regret} how \textbf{suboptimal} the performance of the risk-averse planner is due to being unsure about that function.\par
Also, for the unsafe and the risk-averse planner, I multiplied the feature expectations with a big set of weights sampled from the probability distribution and measured the \textbf{variance} of the resulting set of rewards, called \textit{test variance} for the unsafe planner and \textit{risk-averse variance} for the risk-averse planner. These quantify the aspect of \textbf{uncertainty} the actions of the agent have when ignoring safety, or when trying to take the least risky actions possible.\par
For each one of the above metrics, I took its average over all the test environments, while I also plotted the \textbf{trajectories} of the unsafe and risk-averse planners in each environment of the batch, after finishing the query process for it.
\paragraph{Number of inferences.}
\label{sec:par_num_inf}
As mentioned at the start of \nameref{sec:sec_eval}, an important part of my work was improving the speed of the AIRD \shortcite{mindermannActiveInverseReward2019}, and as the most computationally expensive aspect of that process is Bayesian inference, I used that as the key metric that represents the time progress of the agent's belief about the reward function. Specifically, it is defined as the $\textit{number of Bayesian inferences}=\textit{number of batches}\cdot\textit{number of queries for each batch}\cdot\textit{number of environments in each batch}$, and it is the $X$-axis of the graphs presented in \nameref{sec:sec_results}.
\paragraph{Experiments.}
\label{sec:par_experiments}
First of all, I ran an experiment \textbf{using the AIRD approach}, and the same query selection method that I used in RBAIRD, so as to have data available for comparison and understanding of the aspects where my method improved over it. Additionally, in order to examine the advantages and disadvantages of various parts of RBAIRD, find the parameters that demonstrate optimal performance in each part, and gather enough data to form \textbf{well-informed conclusions}, I performed many experiments where each time I changed one set of parameters while keeping the others \textbf{fixed}. This way, I mitigated the influence of unnecessary parameters and only focused on the ones that matter. The parameters that I varied in each experiment are the following:
\begin{itemize}
  \item The number of \textbf{batches}.
  \item The number of \textbf{environments} in each batch.
  \item The number of \textbf{queries} for each batch.
  \item The method and constants used for \textbf{risk-averse planning}, which were one of the following (described in \nameref{sec:sec_risk_averse}):
    \begin{itemize}
      \item Subtracting the variance from the average reward, with coefficient 1.
      \item Subtracting the variance with coefficient 100.
      \item Taking the worst-case, per state, reward from a set of 10 weight samples.
      \item Taking the worst-case reward with 100 samples.
    \end{itemize}
\end{itemize}
Specifically, regarding the first three parameters, the graphs in \nameref{sec:sec_results} refer to (4 batches, 5 queries, 5 environments) as ``\textit{basic RBAIRD}'', (4 batches, 3 queries, 10 environments) as ``\textit{big batches}'', and (11 batches, 1 query, 5 environments) as ``\textit{many batches}''. Also, ``\textit{subtracting with low coefficient}'' means subtracting the variance with coefficient 1, while \textit{high coefficient} means a coefficient of 100.
\subparagraph{Adding new features.}
\label{sec:sec_new_feat}
Also, as a primary objective of RBAIRD is to be able to adapt to new features in unknown environments, I performed an experiment where I \textbf{progressively added new features} to the training environments. Specifically, in every new batch, I took a set of features that were set to be 0 in all the environments of the previous batches and allowed them to take some value in the new batch (while creating the new environments). For example, if I had 5 batches and 10 features, the environments of the first batch would have the 8 last features set to 0, the second batch only the last 6 features, \ldots, and the last one would have all the features available. I performed the experiment with the following parameters: (6 batches, 2 queries, 5 environments). This experiment simulates the fact that in every new batch the agent can learn about new features that were previously unknown, and still \textbf{adapt and learn} the reward function weight for them.
\section{Results}
\label{sec:sec_results}
\begin{figure}[h]
  \begin{subfigure}[t]{0.47\textwidth}
    \centering
    \includegraphics[width=\textwidth]{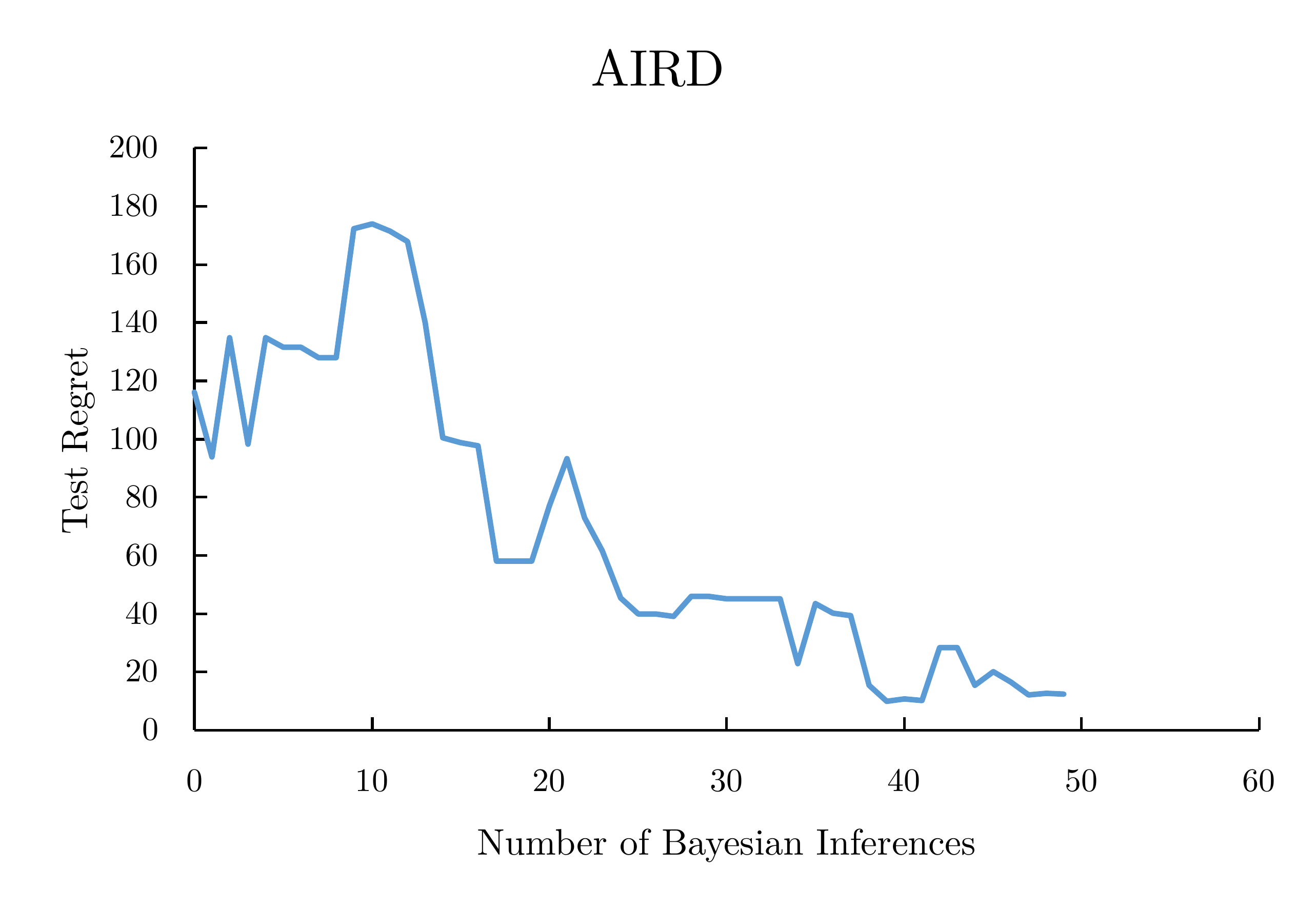}
    \caption{A graph plotting the test regret of AIRD after multiple queries that are selected randomly and then optimized.}
    \label{fig:graph_AIRD_test_regret}
  \end{subfigure}
  \hfill
  \begin{subfigure}[t]{0.47\textwidth}
    \centering
    \includegraphics[width=\textwidth]{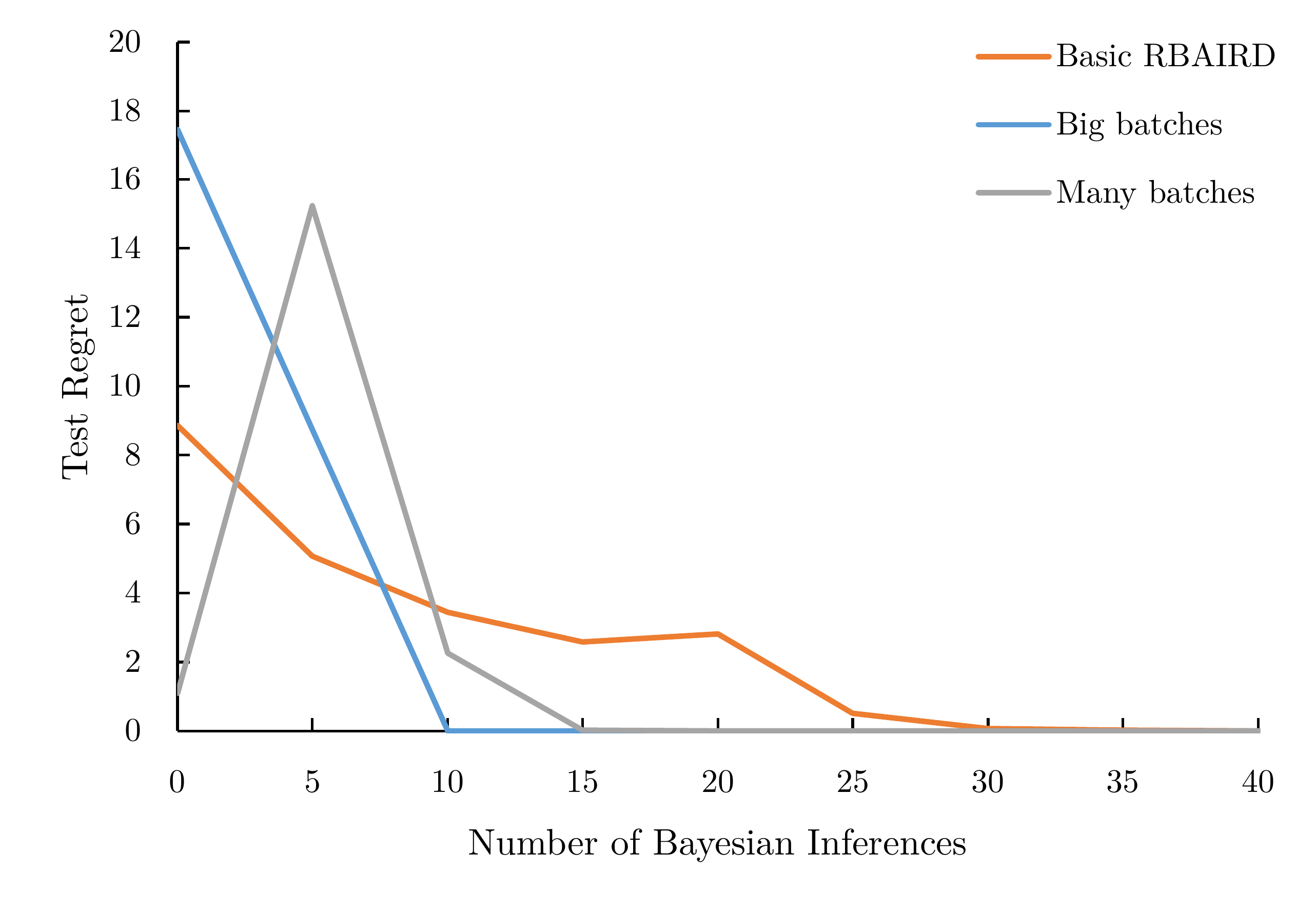}
    \caption{A graph plotting the test regret of RBAIRD using the same query selection method, in experiments with three different configurations of (number of batches, batch size, number of queries), described in \nameref{sec:par_experiments}.}
    \label{fig:graph_rbaird_test_regret}
  \end{subfigure}
  \caption{A comparison of the ability of AIRD and RBAIRD to understand the reward function, in terms of both accuracy and speed (regarding the number of inferences used.)}
	\label{fig:aird_rbaird_comparison}
\end{figure}

Now I will present the results of the experiments described in \nameref{sec:par_experiments}. In general, they validate the hypotheses made in \nameref{sec:sec_intro}, by showing a significant improvement in performance and a smaller need for human intervention, better accuracy, adaptability to new environments, and reduction in the uncertainty of the actions taken by the agent.
\paragraph{AIRD performance.}
\label{sec:par_AIRD_results}
For comparison purposes, we will have a look at the performance of the AIRD \shortcite{mindermannActiveInverseReward2019} approach, when using \textit{discrete queries} chosen randomly and then optimized for maximal information gain, which is the same method that I used for my work. As shown in \Fref{fig:graph_AIRD_test_regret}, the performance approaches optimal after about $50$ queries, but never finds the true reward function completely, as a single environment often isn't enough to capture all the different features of it.\par
It should be noted that other query selection methods in the AIRD paper perform better, but are very computationally expensive, thus I couldn't measure their performance for many queries. However, as the results are comparative, and the method used is the same, it is assumed that they should still hold in other selection methods, which could be tested as described in \nameref{sec:sec_lim_future_work}.
\subsection{Batch Queries}
We will now examine the capability of RBAIRD in improving the above performance. For that purpose, in \Fref{fig:graph_rbaird_test_regret}, I present the graph of the \textit{test regret} of my method, as it is directly linked to the \textbf{accuracy of the agent's belief} about the reward function. I plot that metric depending on the number of \textbf{Bayesian inferences} (described in \nameref{sec:par_num_inf}), in three experiments with different number of batches, number of environments in each batch, and number of query iterations for each batch, as analyzed in \nameref{sec:par_experiments}. This way, I measure the impact batches have on RBAIRD's performance, and which size parameters work the best.\par
I observed that in all the experiments the process exactly found the reward function, after $\mathbf{30}$ \textbf{inferences} (6 queries) when having 4 batches, 5 environments in each batch, and 5 queries for each batch, $\mathbf{15}$ \textbf{inferences} (3 queries) with 11 batches and 1 query and $\mathbf{10}$ \textbf{inferences} (1 query) when having 10 environments in each batch and 3 queries. First of all, we see a significant improvement in performance over the AIRD approach. This can be explained by the fact that when we examine the effects of a reward function over multiple environments, we get much more information about the behavior it incentivizes in \textbf{different situations}. This cannot be achieved as efficiently by answering different queries in the same environment. The fact that the experiments with more or bigger batches, therefore fewer queries in each environment, performed at least \textbf{2 times better} than the \textit{basic RBAIRD} experiment, also supports our previous hypothesis: asking the same query over multiple environments or many queries in different environments instead of the same one provides much more \textbf{diverse information} about the reward function we need to find, and therefore \textbf{improves the performance significantly}.\par
In the real world, the trade-off between the size of batches and the number of them can be determined by the computational cost of inference, query, and training of the agent on the updated reward probabilities, as well as the available human resources needed for providing answers to the process's queries (based on the number of them).
\subsection{Risk-averse Planning}
\begin{figure}[tb]
  \begin{subfigure}[t]{0.47\textwidth}
    \centering
    \includegraphics[width=\textwidth]{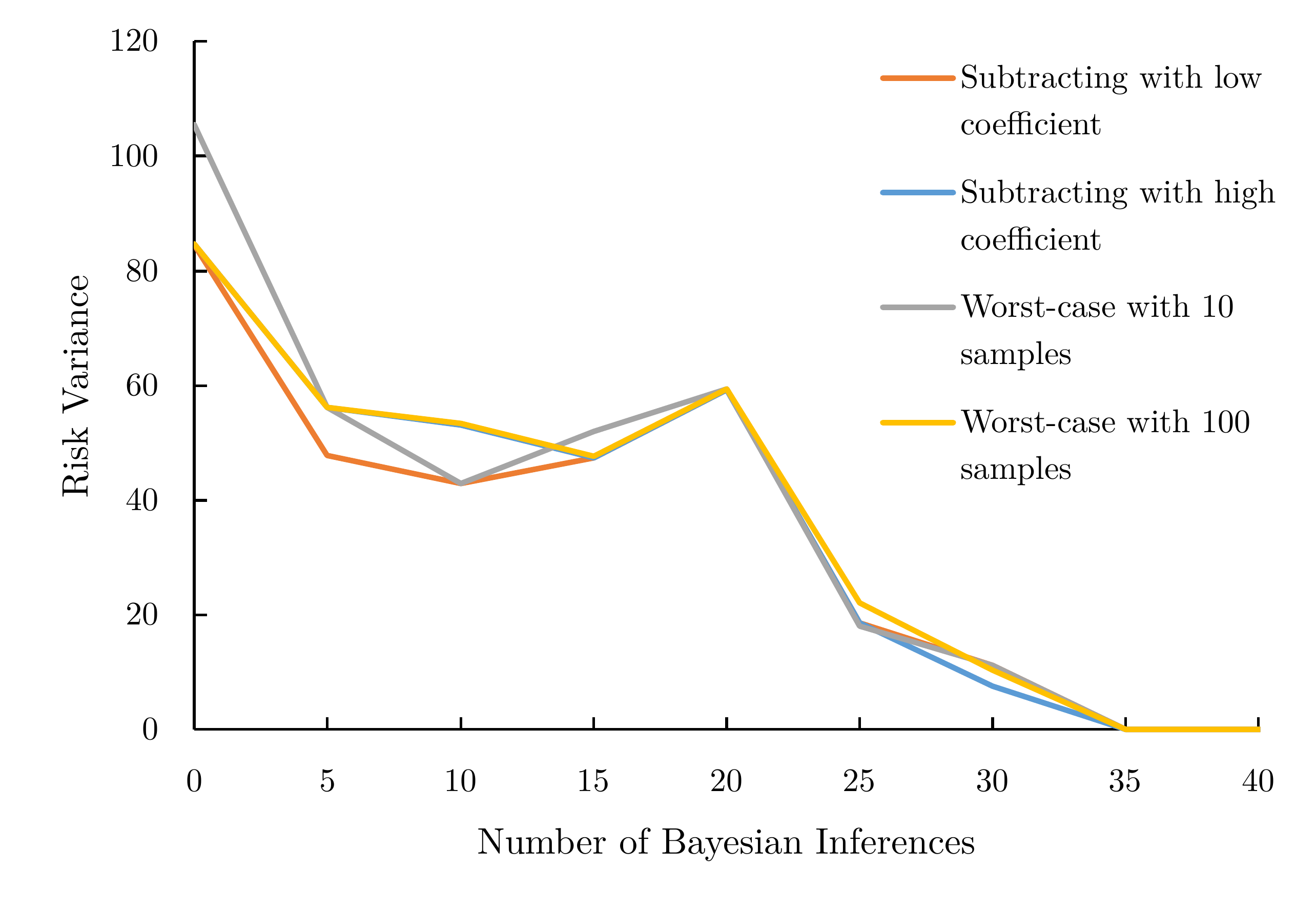}
    \caption{A plot of the risk-averse variance using different risk-averse planning methods.}
    \label{fig:graph_risk_methods_risk_variance}
  \end{subfigure}
  \hfill
  \begin{subfigure}[t]{0.47\textwidth}
    \centering
    \includegraphics[width=\textwidth]{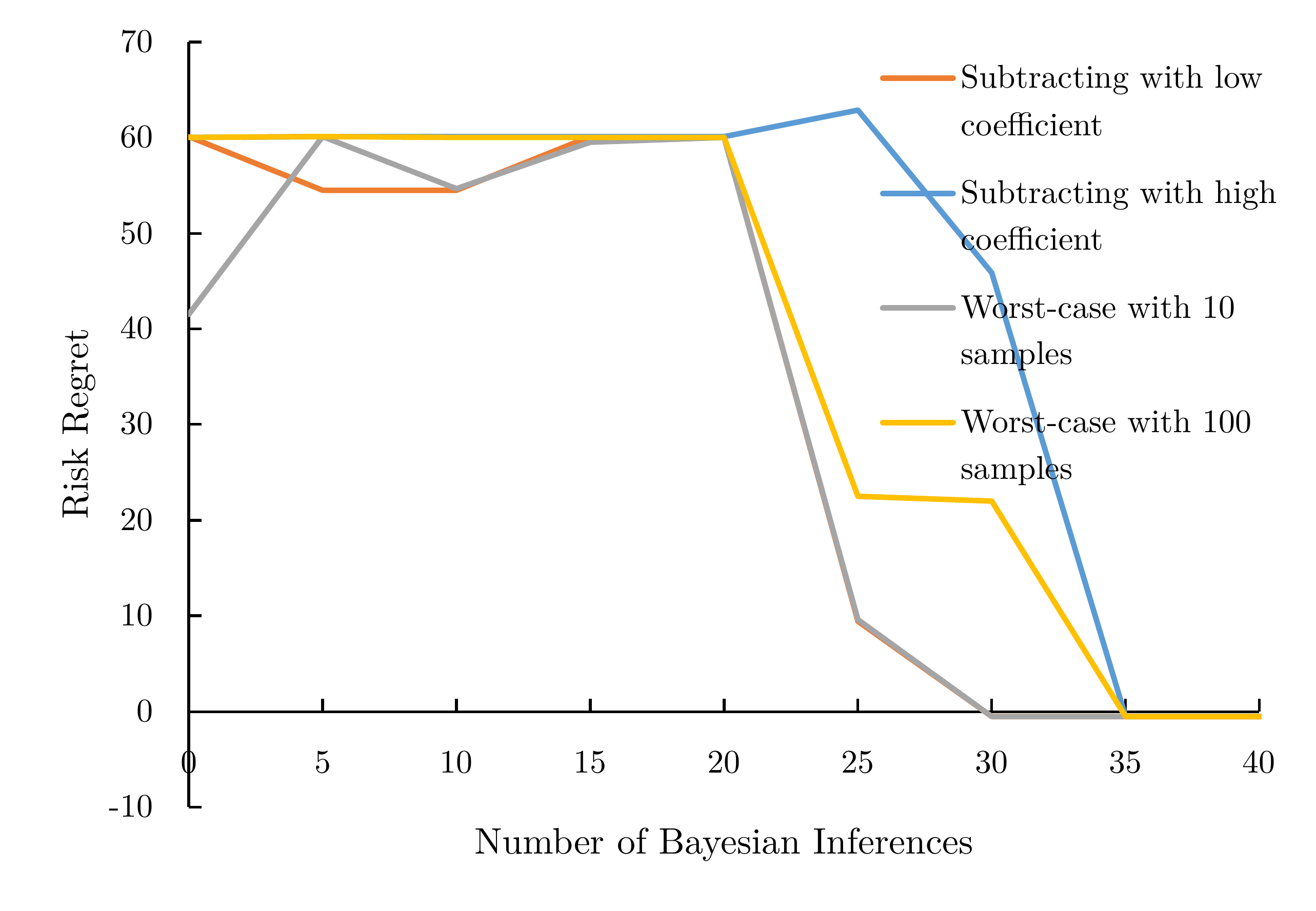}
    \caption{A plot of the risk-averse regret using different risk-averse planning methods.}
    \label{fig:graph_risk_methods_risk_regret}
  \end{subfigure}

  \begin{subfigure}[t]{0.47\textwidth}
    \centering
    \includegraphics[width=\textwidth]{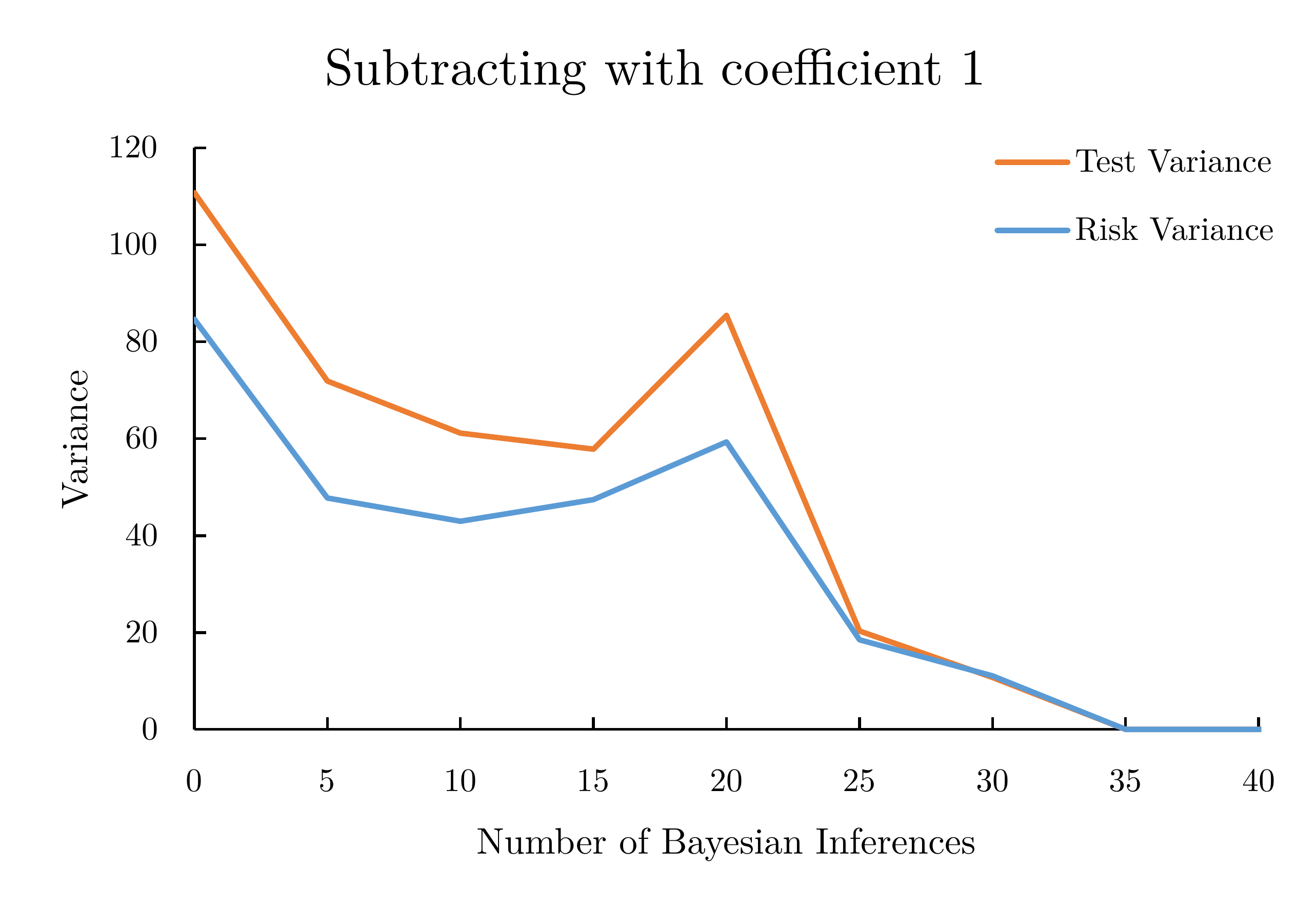}
    \caption{A plot comparing the test variance with the risk-averse variance, when the risk-averse planner is penalized by subtracting the variance with coefficient 1.}
    \label{fig:graph_rbaird_test_vs_risk_variance}
  \end{subfigure}
  \hfill
  \begin{subfigure}[t]{0.47\textwidth}
    \centering
    \includegraphics[width=\textwidth]{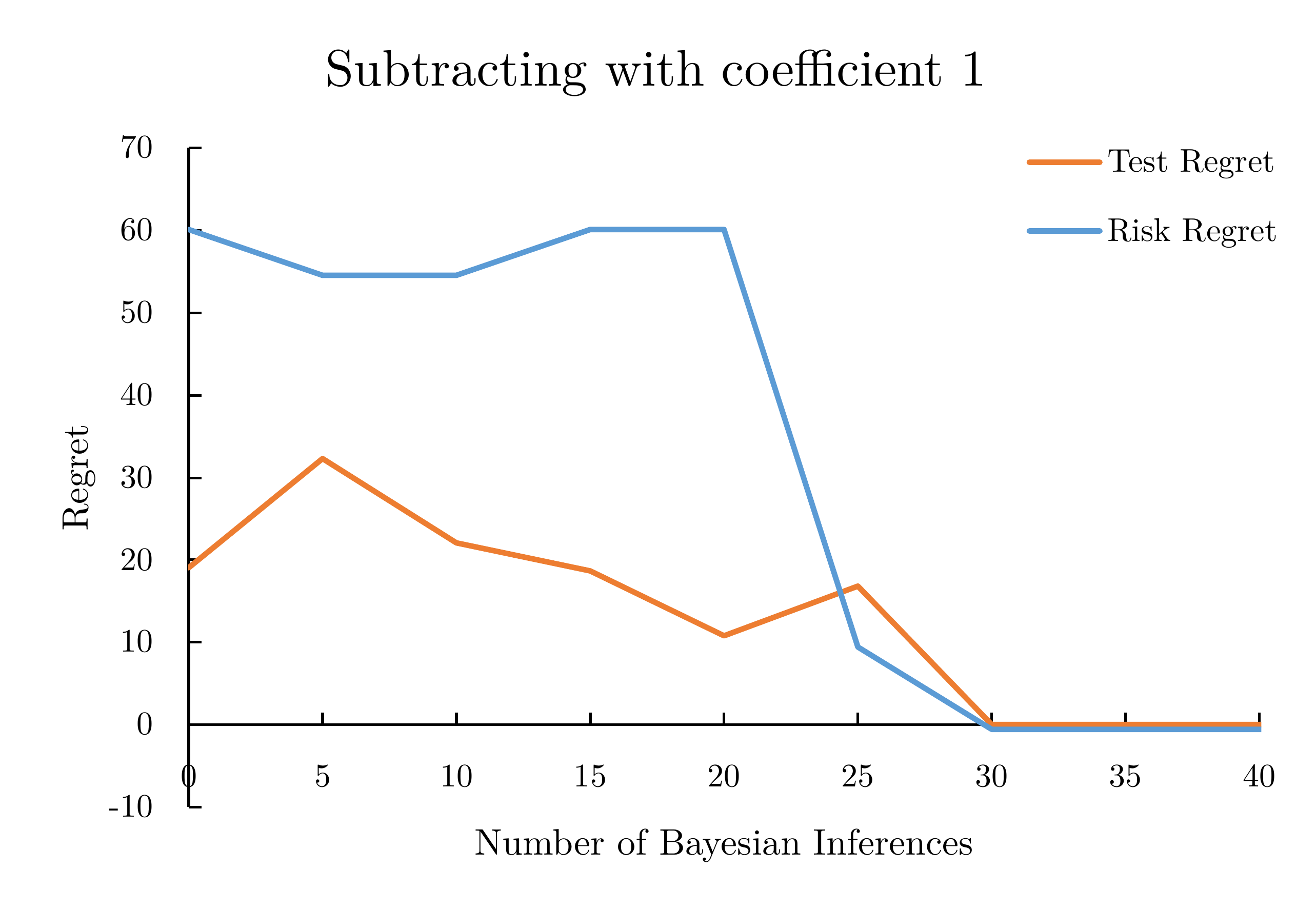}
    \caption{A plot comparing the test regret with the risk-averse regret, when the risk-averse planner is penalized by subtracting the variance with coefficient 1.}
    \label{fig:graph_rbaird_test_vs_risk_regret}
  \end{subfigure}
  \caption{A comparison of the performance of different risk-averse planning methods, and a demonstration of the effect that subtracting the variance of the planner's actions has on the suboptimality of the total reward of a trajectory in the environment and the uncertainty of that reward at the time.}
	\label{fig:rbaird_test_vs_risk}
\end{figure}
\begin{figure}[tb]
  \begin{subfigure}[t]{0.47\textwidth}
    \centering
    \includegraphics[width=\textwidth]{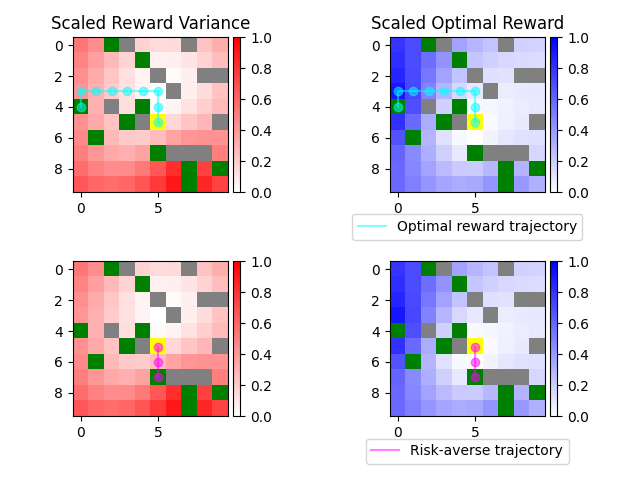}
    \caption{A visualization of the trajectories when the risk-averse planning method takes the worst-case scenario of 100 sampled rewards.}
    \label{fig:traj_worst_100}
  \end{subfigure}
  \hfill
  \begin{subfigure}[t]{0.47\textwidth}
    \centering
    \includegraphics[width=\textwidth]{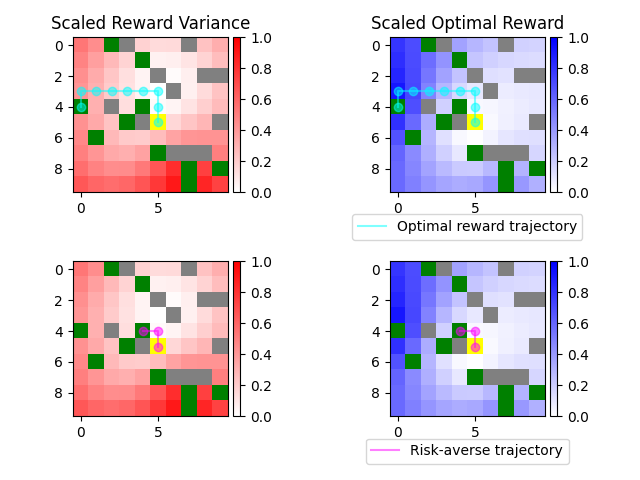}
    \caption{A visualization of the trajectories when the risk-averse planning method subtracts the variance of each state with a coefficient of 100.}
    \label{fig:traj_sub_100}
  \end{subfigure}
  \caption{A comparison of the trajectories of the unsafe and risk-averse planner computed using different risk-averse planning methods in the same environment. The cell's color indicates the relative optimal reward in each state (blue) and the variance of it based on the probability distribution (red), where higher color intensity means a bigger value of the respective measure. Green cells are goals, gray cells are walls, and the yellow cell is the start.}
	\label{fig:risk_comp_traj}
\end{figure}

The second key part of RBAIRD is \textbf{risk-averse planning}, which penalizes the uncertainty of the agent's actions in order to incentivize safer behavior, as described in \nameref{sec:sec_risk_averse}. This can be accomplished by multiple methods, some simpler and some more sophisticated, with a key \textbf{trade-off between reward}(\textit{risk regret}) \textbf{and certainty}(\textit{risk variance}).
\paragraph{Method comparison.} I examined two simple methods: subtracting the variance of the state reward and taking the worst-case reward, and tested them with two different parameter values for each one (variance coefficient and number of weight samples respectively). As demonstrated in \Cref{fig:graph_risk_methods_risk_regret,fig:graph_risk_methods_risk_variance}, between the experiments I ran, the one that demonstrated the best comparative performance, both regarding risk regret and risk variance, was \textbf{subtracting the variance with coefficient 1} from the expected reward (based on the probabilities computed at the time). It also showed the fastest \textbf{convergence to the optimal policy}, even when still being unsure about the reward function.
\paragraph{Risk-averse vs unsafe performance.} When comparing the performance of that risk-averse planning method with that of the unsafe planner, like in \Cref{fig:graph_rbaird_test_vs_risk_regret,fig:graph_rbaird_test_vs_risk_variance}, we observe that the risk-averse planner takes \textbf{consistently more certain actions} than the unsafe one. However, when the agent is still unsure about the true reward function and the influence of his actions' variance is bigger, they are much more conservative, by taking very safe routes that lead to much smaller reward. On the other hand, that suboptimality of the agent's safe trajectories abruptly diminishes \textbf{at the same time the unsafe performance becomes optimal}, which is when the agent is almost sure about the intended behavior. Also, more sophisticated approaches, like those described in \nameref{sec:sec_lim_future_work}, can \textbf{significantly improve that performance} and reduce that big loss in total reward for the sake of lower variance, which often leads to the ignorance of sufficiently safe but also highly rewarded trajectories, a phenomenon commonly referred to as \textit{blindness to success}.
\paragraph{Trajectory comparison.} In \Fref{fig:risk_comp_traj}, I present some images that represent the trajectories of the robot in the same environment, when using the risk-averse planner and the unsafe planner, and two different risk-averse planning methods. As they are enhanced with visual data about the \textit{reward} at each specific state (which is the intensity of the blue color of the cells in the second column of the figure grids) and the \textit{variance} of the states' reward (the intensity of the red color in the first column), we could possibly infer the \textbf{strategy of the respective planners} and methods used, while evaluating the extent to which the agent values the variance more than the reward. We can observe that both safe planning methods made the agent prefer to stay close to the start state because of the high cost of the variance and the low gain because of the reward, but they chose different routes because of the different ways they value certainty.
\subsection{Adaptability to new features}
\begin{figure}[tb]
  \begin{subfigure}[t]{0.47\textwidth}
    \centering
    \includegraphics[width=\textwidth]{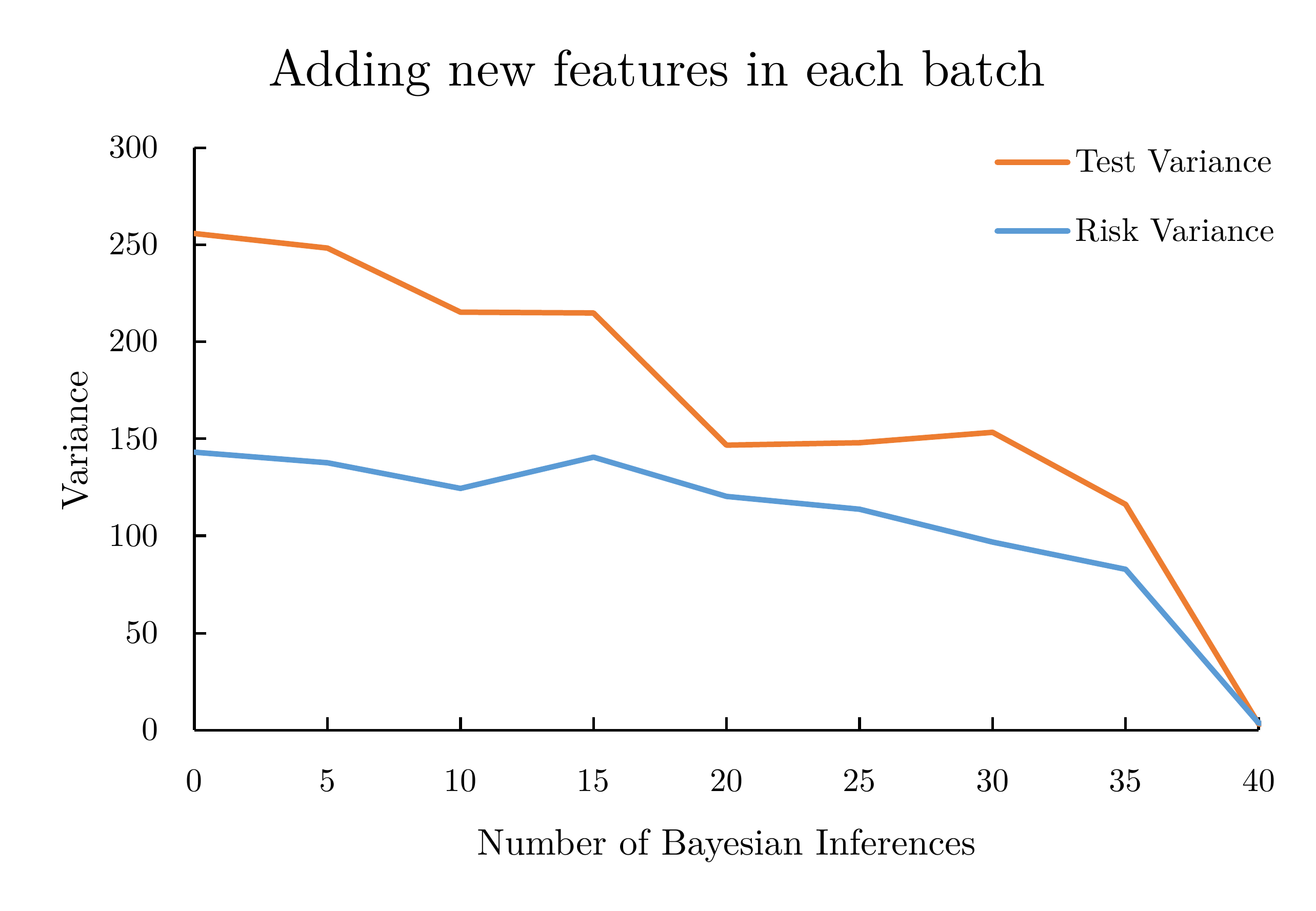}
    \caption{A plot showing the evolution of the risk variance, in comparison with the respective test variance, when continuously adding new features in the environments of each batch.}
    \label{fig:graph_new_feat_test_vs_risk_var}
  \end{subfigure}
  \hfill
  \begin{subfigure}[t]{0.47\textwidth}
    \centering
    \includegraphics[width=\textwidth]{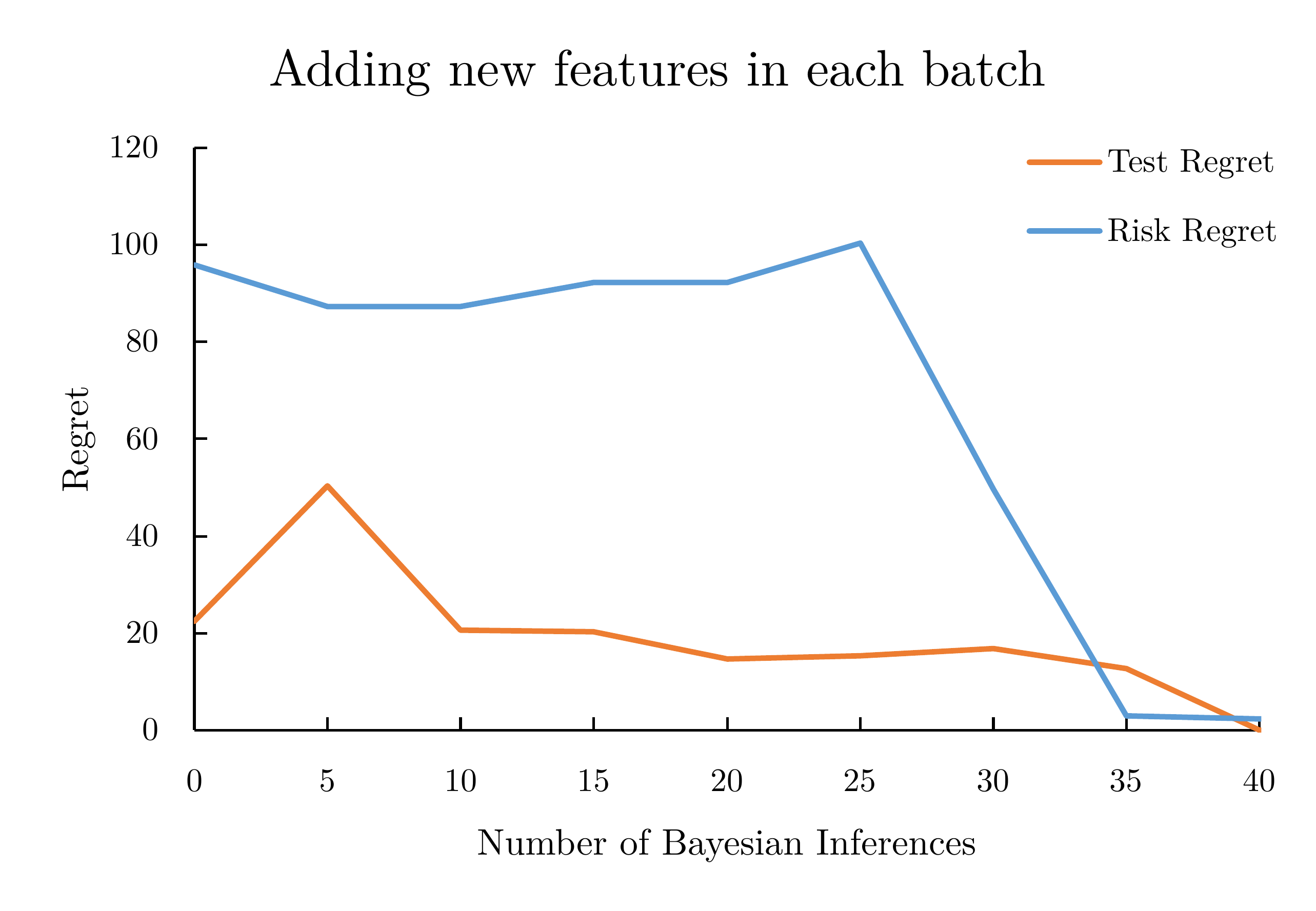}
    \caption{A plot showing the evolution of the test regret and the risk regret, when continuously adding new features in the environments of each batch.}
    \label{fig:graph_new_feat_test_vs_risk_reg}
  \end{subfigure}
  \caption{A demonstration of RBAIRD's performance, in terms of uncertainty and reward regret, when continuously encountering new features in unknown environments, like most real-world scenarios.}
	\label{fig:new_features}
\end{figure}

The results of the experiment described in \nameref{sec:sec_new_feat} are a great indication of the RBAIRD's ability to perform in the real world, where new environments often contain \textbf{unknown features} and the agent needs to learn how to treat them. As shown in \Fref{fig:new_features}, the number of Bayesian inferences needed to reach optimal performance, both when using the risk-averse and the unsafe planner, is about 40, lower than AIRD, and only a little bit higher than when it had all the features available at the start, where we needed at most 30 queries, as shown in \Fref{fig:graph_rbaird_test_regret}. In the experiment, I added only 2 queries per batch. These results show that RBAIRD is able to \textbf{adapt to unforeseen features} very quickly, and learn new aspects of the reward function needed to obtain the intended behavior in real-life scenarios instead of training. Also, the risk-averse variance was almost half the test variance when the agent was very uncertain about the reward function, noting the importance of the risk-averse planner when encountering unknown environments and the \textbf{safety it offers in these situations}.\par
AIRD didn't have the aforementioned capabilities, since it only involved training on a single environment and repeating the query process until it captured every aspect of it that can be distinguished in that one environment. It also didn't have any safety measures in place for the scenarios when new features appear, when the agent made risky decisions, aiming for the highest expected reward but ignoring the penalty that the unknown features could cause.
\section{Related work}
\label{sec:sec_related_work}
There are various approaches related to this work, closely relevant to \textit{Inverse Reinforcement Learning} \shortcite{ngAlgorithmsInverseReinforcement2000,aroraSurveyInverseReinforcement2020}, and most of them belong to the field of \textit{Reinforcement Learning from Human Feedback} \shortcite{casperOpenProblemsFundamental2023}. Some of them are similar in the fact that they use a human's preferences over sets (or pairs) of trajectories \shortcite{christianoDeepReinforcementLearning2023}, or even combine them with expert demonstrations \shortcite{ibarzRewardLearningHuman2018} to compute the optimal policy without knowing the reward function. \textit{Cooperative Inverse Reinforcement Learning} \shortcite{hadfield-menellCooperativeInverseReinforcement2016} follows a similar route but is more bidirectional than the others, by making both the human and the agent take actions on the same MDP, where the agent doesn't know the reward function but tries to maximize the total reward. This incentivizes the human to demonstrate teaching behavior instead of just the optimal policy, and the agent to actively learn, in order to find the reward function. On the other hand, \shortcite{aminRepeatedInverseReinforcement2017} tries to mitigate the unexpected behaviors an agent presents when completing several tasks, by providing a human demonstration each time this happens.\par
There is another set of methods that try to ensure safety by mitigating unwanted side-effects caused by the agent's effort to acquire the maximum reward possible. For example, \shortcite{krakovnaPenalizingSideEffects2019} measures the reachability of the current state from a \textit{baseline}, from where any deviation in aspects of the environment different than the ones we want will be penalized. Also, \shortcite{panEffectsRewardMisspecification2022} examines reward hacking and tries to increase our control over unpredicted behaviors of AI systems, by detecting thresholds where that behavior abruptly changes. Finally, \shortcite{shahPreferencesImplicitState2018} exploits the fact that our world and behavior are like an optimized MDP, to the extent that the world's state is caused by human behaviors that follow the same rules and constraints that we want the agent to follow in order to not cause side-effects when pursuing a task. These world states in certain environments can be combined with an IRL algorithm to infer the correct reward function that penalizes the robot when altering the environment's state in aspects we don't want it to.
\section{Limitations and Future work}
\label{sec:sec_lim_future_work}
\paragraph{Query selection methods.} In this work I only used a discrete query selection method, from those that AIRD used, that is based on selecting 5 random reward functions one by one and optimizing them, using gradient descent, to increase the information gain, as described in \nameref{sec:par_query_selection_aird}. This was done because of the computational cost of the other methods (and query sizes), and my limited computing power resources, which caused me to be unable to run them. However, these methods demonstrate much better and faster inference about the reward function, as shown in the AIRD paper, they could greatly improve the performance of RBAIRD, and pave the way for tackling more complex environments, similar to those used in the real world. Therefore, a study on these query selection methods (and possibly other ones that are more efficient), and an effort to reduce their computational complexity, in order for them to be more usable, would be very significant.
\paragraph{Blindness to success.} The risk-averse reward functions used in my work are very simple ones, and have big flaws in their performance. The most important of them is a phenomenon called \textit{blindness to success}, which means that they value uncertainty much more than the reward itself, causing them to even ignore their specific goal and just stay in states that are considered safe. An even more extreme situation would be that the variance penalty coefficient would be so high that the agent decides to perform a harmful action, if it is absolutely certain about the reward it will get because of it (and any intended action has high uncertainty). There are some more sophisticated and complex approaches to this issue, like \shortcite{greenbergEfficientRiskAverseReinforcement2022}, that focus on implementing risk-averse planning in a way that it remains usable and has high performance, while still considering edge-case scenarios that are important in certain situations and use cases.
\paragraph{Query answer for the whole batch.} Currently, RBAIRD computes a single query for the whole batch, but the human needs to answer it for each environment in the batch separately. This causes the number of answers to be the $\textit{number of queries}\cdot\textit{batch size}$, largely increasing the human intervention needed. What could be done is to make the human answer the query by selecting the reward function that performs the best in all environments of the batch (considering overall performance), reducing the number of answers the human needs to give. However, this has some significant flaws. First of all, an answer of that type is very ambiguous, as there could be reward functions that perform the best in some environments and the worst in others, making human's work much harder in judging the reward functions. There isn't some clear measure that can be combined over all the environments so that the human can give one answer. Also, after running some experiments where I tried to come up with such a measure, like the sum of the rewards of a planner trained using the same reward function over all environments, I observed that the RBAIRD process then doesn't work at all, as the probability distribution doesn't converge to a single reward function, but it stays uncertain.
\paragraph{Human query answering and metrics.} The way I evaluated the performance of RBAIRD (and the way AIRD did it as well) was by using a process that knew the true reward function, used each reward function of the query to compute their feature expectations on the desired environment, and computed the true reward that the planner would get in the real world. It then used these rewards to compute a probability distribution (that represented a rational decision maker) from where it sampled the function which was the answer to the query. This way, it simulated the answer a human would probably give, but it didn't take into account some difficulties that a human would encounter. Specifically, the human doesn't know the true reward function either, and judging a possible reward function just by its weights is very difficult, as he needs to distinguish the behaviors that the function incentivizes in a specific environment. An addition that would make it easier is making an interactive query-answer environment that presents the visualizations of the optimal trajectories derived from the query functions in that environment, as well as other useful metrics except for the total expected reward. This could be enhanced by highlighting certain steps where the trajectory is very different than that of the other functions and noting some higher-level features/patterns or general aspects of the environment (not just the features that the function considers) that cause incentivization of a certain behavior by that function. This could be implemented using a Machine Learning algorithm that is trained on the trajectories of the agent and provides the wanted characteristics, but this is related to interpretability \shortcite{g.j.rudnerKeyConceptsAI2021,christianoElicitingLatentKnowledge2021}. That query answer environment could also provide the unsafe and the risk-averse behavior caused by the probabilities computed in some steps of the query process, as a safety measure to prevent the emergence of malicious or harmful behaviors.
\section*{Acknowledgments}
I would like to thank Peter McIntyre and the Non-Trivial Fellowship, who helped and guided me through the process of solving the world's most pressing problems, leading me to be intrigued by AI Alignment, come up with, and refine the idea of RBAIRD. I would also like to thank S{\"o}ren Mindermann, who evaluated and supported my idea, and gave me access to the code of Active Inverse Reward Design, which I used and modified to implement the work presented here.
\bibliographystyle{apacite}  
\bibliography{references}  
\end{document}